\documentclass[letterpaper]{article}
\usepackage{aaai}
\usepackage{times}
\usepackage{helvet}
\usepackage{courier}
\usepackage{lettrine}
\usepackage{amsmath}
\usepackage{xcolor}
\usepackage{graphicx}
\usepackage{amssymb}
\usepackage{subfig}
\usepackage{booktabs}
\usepackage{relsize}
\usepackage{textcomp}
\usepackage{hyperref}
\usepackage{commath}
\usepackage{algorithm}
\usepackage{algorithmic}
\usepackage[nameinlink,noabbrev]{cleveref}
\usepackage{array}
\newcolumntype{P}[1]{>{\centering\arraybackslash}p{#1}}

\hypersetup{%
  colorlinks=true,
  linkcolor=blue,
  linkbordercolor=red,
  citecolor=blue,
  linkcolor = blue,
  anchorcolor = blue,
  filecolor = blue,
  urlcolor = red
}

\begin{document}
%
\title{Simultaneous Identification of Tweet Purpose and Position}
\author{Rahul Radhakrishnan Iyer\thanks{Corresponding Author: \texttt{rahuli@alumni.cmu.edu}}\\
Language Technologies Institute\\
Carnegie Mellon University\\
Pittsburgh PA 15213\\
\And
Yulong Pei\\
Robotics Institute\\
Carnegie Mellon University\\
Pittsburgh PA 15213\\
\And
Katia Sycara\\
Robotics Institute\\
Carnegie Mellon University\\
Pittsburgh PA 15213\\
}

 
\maketitle
\begin{abstract}
\begin{quote}
Tweet classification has attracted considerable attention recently. Most of the existing work on tweet classification focuses on topic classification, which classifies tweets into several predefined categories, and sentiment classification, which classifies tweets into positive, negative and neutral. Since tweets are different from conventional text in that they generally are of limited length and contain informal, irregular or new words, so it is difficult to determine user intention to publish a tweet and user attitude towards certain topic. In this paper, we aim to simultaneously classify tweet purpose, i.e., the intention for user to publish a tweet, and position, i.e., supporting, opposing or being neutral to a given topic. By transforming this problem to a multi-label classification problem, a multi-label classification method with post-processing is proposed. Experiments on real-world data sets demonstrate the effectiveness of this method and the results outperform the individual classification methods.\\~\\

\-\hspace{0.5cm} \textbf{\textit{key words ---}} multi-label classification, tweet purpose classification, tweet position classification, natural language processing, machine learning
\end{quote}
\end{abstract}

\section{Introduction}
\label{sec:intro}
Over the past few years, microblogs have become one of the most popular online social networks. Twitter and Weibo are two of the most representative microblog platforms. With more than 400 million tweets per day on Twitter and more than 66 million daily active users on Weibo, microblog users generate large amount of tweets which cover rich topics including political issues, celebrity gossip, or personal life. Because the user generated content (UGC) on microblogs covers rich topics and is real-time, mining and analyzing this information is beneficial to both industrial community and academic community. Since human inspection of this vast stream of real-time data is expensive and time-consuming, automatic computational tools and mining methods are thus in huge demand. Since tweets are different from conventional text and they generally are of limited length and contain informal, irregular or new words, it is difficult to exploit user intention to publish a tweet and attitude towards certain topic only from the analysis on topic and sentiment level. Recently, a variety of studies have been done on Twitter including event detection, user recommendation and tweet classification. Tweet classification attracts considerable attention since it is very important to analyze, understand and predict user behaviors on social networks.

Most of the existing work focuses on either tweet purpose classification or position classification. For purpose classification, previous work \cite{naaman2010really,alhadi2011exploring} has classified tweets into different classes of purpose, e.g., social interaction with people, promotion or marketing, information sharing, etc. For position classification \cite{saif2012alleviating,go2009twitter,kouloumpis2011twitter}, tweets are classified into positive, negative and neutral. However, dealing with tweet purpose and position separately in previous work has two limitations. First, in order to determine the purpose and position of a tweet, two different classifiers should be trained and this is inefficient. Second, the correlation between the tweet purpose and the position has not been exploited. For example, given the topic Obama care, based on the data set introduced in Table \ref{table:obama_stats}, when people try to share information, they tend to be positive to this policy, while users are more likely to oppose it when they interact with others. If these correlations are captured, it will be beneficial for tweet position and purpose classification.

In order to overcome these limitations, in this study we aim to identify tweet purpose and position simultaneously by exploiting the correlation between purpose and position in tweets. Tweet purpose indicates user’s intention in publishing a tweet, such as sharing information or expressing personal emotion. Tweet position indicates whether user will support, oppose or be neutral to a given topic. We transform this problem to identify tweet purpose and position simultaneously into a multi-label classification problem. Our method is advantageous in two aspects: ($1$) It is more efficient to use multi-label classification methods to simultaneously identify tweet purpose and position since only one unified classifier needs to be trained. ($2$) The correlation between tweet purpose and position can be captured by multi-label classification methods to improve the accuracy for classification. Besides, aiming to tackle the issue that some tweets in the data are predicted to contain no labels or multiple labels using multi-label classification method, two different post-processing strategies have been proposed. In order to validate the effectiveness of this problem transformation and post-processing strategies, we build two data sets collected from Twitter and experiments are conducted on the data sets.

In short, this paper makes the following contributions:
\begin{itemize}
    \item We define the task to identify tweet purpose and position simultaneously and transform this problem to a multi-label classification problem.
    \item We propose two post-processing strategies i.e., summation and weighted summation, for the classification task and by incorporating the strategies into the multi-label classification method, the classification performance can be improved.
    \item We test our approach on two real-world data sets to validate the classification method with post-processing and the results demonstrate the effectiveness of the problem transformation and post-processing strategies.
\end{itemize}

The rest of this paper is organized as follows. Section \ref{sec:related} reviews the related work. Section \ref{sec:data} presents the collection and annotation of data sets. In Section \ref{sec:multi_label}, the multi-label classification method and the post-processing strategies are introduced. Section \ref{sec:experiments} presents the experiments and finally we conclude this study in Section \ref{sec:conclusions}.

\section{Related Work}
\label{sec:related}
In this section, we review the related work in two areas: sentiment analysis on microblog and tweet purpose identification.

\subsection{Sentiment Analysis on Microblog}
\label{subsec:microblog}
Sentiment analysis has been playing a crucial role in natural language processing and text mining recently and it aims to analyze people’s opinions, sentiments, evaluations, appraisals, attitudes, and emotions towards entities such as products, services, organizations, individuals, issues, events, topics, and their attributes \cite{liu2012sentiment}. Most of existing work aims at sentiment polarity classification, i.e., classifying opinion into $2$ categories (positive and negative) or $3$ categories (positive, negative and neural) and in general, they conducted experiments on the movie or product review data.

With the popularity of microblog, some sentiment analysis work on microblog platform has been done. Due to the short length of tweets, the use of informal and irregular words, and the rapid evolution of language \cite{saif2012alleviating} on the Internet, sentiment analysis on microblog is more challenging than conventional sentiment analysis on reviews or weblog posts. Go et al. \cite{go2009twitter} applied a distant supervised learning method to classify tweet sentiment, and emoticons have been used as noisy labels for training data. Kouloumpis et al. \cite{kouloumpis2011twitter} exploited hashtags in tweets to build training data. Their experiments demonstrated the part-of-speech features may not be useful in classifying sentiment on tweets. Apart from supervised methods, unsupervised methods have also been used in sentiment analysis on tweets. He et al. \cite{he2012quantising} proposed a probabilistic generative model for joint modeling sentiment labels and topics for tweets. Recently, several approaches involving natural language processing \cite{iyer2019event,iyer2019unsupervised,iyer2019heterogeneous,iyer2017detecting,iyer2019machine,iyer2017recomob}, machine learning \cite{li2016joint,iyer2016content,honke2018photorealistic}, deep learning \cite{iyer2018transparency,li2018object} and numerical optimizations \cite{radhakrishnan2016multiple,iyer2012optimal,qian2014parallel,gupta2016analysis,radhakrishnan2018new} have also been used in the visual and language domains.

\subsection{Tweet Purpose Identification}
\label{subsec:tweet_purpose}
Users’ intentions for using microblogs have been widely studied recently \cite{naaman2010really,alhadi2011exploring,mohammad2013identifying}. Most studies consider tweet purpose identification as a classification task. Naanman et al. \cite{naaman2010really}, from the perspective of characteristics of social activity and communication patterns on Twitter, categorized tweets into $9$ types: information sharing, self promotion, opinions/complaints, statements and random thoughts, me now, question to followers, presence maintenance, anecdote (me) and anecdote (others). Alhadi et al. \cite{alhadi2011exploring} organized tweets purpose to taxonomy which includes social interaction with people, promotion or marketing, sharing of resources, giving/requiring feedback, broadcast alert/urgent information, requiring/raising funding, recruitment of worker, and expression of emotions. Mohammad et al. \cite{mohammad2013identifying} studied tweet purpose on the electoral topic. They firstly organized these political tweets to $3$ types of purposes (favor, oppose, and other) and furthermore classified these $3$ categorizes into $11$ sub-categorizes according to the emotion degrees.

Some studies are more related to our work. In \cite{huang2013sentiment}, both sentiment and topic for tweets have been modeled using a unified framework. However, this work is different from ours because it has not explored the purpose and position for tweets and the classifiers used for sentiment and topic are trained separately using different features (although the title of this paper used the keywords ``multi-label classification approach"), while in our study, only one multi-label classifier will be trained and purpose and position labels can be obtained simultaneously.

\section{Data Set}
\label{sec:data}
In this section, the data collection and the label annotation rules for purpose and position are introduced.

\subsection{Data Collection}
\label{subsec:collection}
In order to build the Twitter data set, we collected the tweets in two topics, i.e., Obama care and death penalty. We used Twitter Search API\footnote{\url{https://dev.twitter.com/docs/using-search}} with the queries Obama care and death penalty. Then we pre-processed these tweets by removing ($1$) non-English tweets, ($2$) tweets less than $5$ words, and ($3$) duplicated tweets. After removing the irrelevant tweets to these two topics, we labeled $1000$ tweets for each of the two topics. The statistics of two data sets are shown in Tables \ref{table:obama_stats} and \ref{table:death_stats} and the details for annotation are introduced in the following subsection.

\subsection{Data Annotation}
\label{subsec:annotation}

\subsubsection{Purpose Label}
\label{subsubsec:purpose_label}
In Section \ref{sec:related}, some studies on tweet purpose classification have been reviewed. Based on previous studies and characteristics of our data sets, we organize tweets into $3$ categorizes: ($1$) Express emotion/personal interests; ($2$) Information sharing; and ($3$) Social interaction. Some example tweets of different purpose labels are shown in Table \ref{table:tweet_purpose}.

\subsubsection{Position Label}
\label{subsubsec:position_label}
Position labels are based on the position in the tweet towards the given topic, i.e., Obama care and death penalty, and consist of three types of labels: pro, con and neutral. Some examples of position labels are shown in Table \ref{table:tweet_position}.

\begin{table*}[ht]
\vspace{2ex}
\resizebox{\textwidth}{!}{\begin{tabular}{ c | P{13cm} }
\toprule
\textbf{Purpose Label} & \textbf{Example of Tweet}\\
\midrule
Express emotion & I looked at Obamacare and said, ``Yeah. And??" I'm not alone in thinking it was a mistake to support and help re-elect him. Embarrassing. \\\specialrule{\cmidrulewidth}{0pt}{0pt}
Information sharing & Jimmy Kimmel found that people support the Affordable Care Act much more than Obamacare \url{http://t.co/eTX46m9ZVi} $\#$PoliticalNews \\\specialrule{\cmidrulewidth}{0pt}{0pt}
Social interaction & $@$DamianBennett $@$SheilaKihne oh, I’m sorry, did Obamacare pass with unanimous support from Republicans? Or the opposite of that? \\
\bottomrule
\end{tabular}}
\centering
\caption{Example tweets of different purpose labels.}
\label{table:tweet_purpose}
\end{table*}

\begin{table*}[ht]
\vspace{2ex}
\resizebox{\textwidth}{!}{\begin{tabular}{ c | P{13cm} }
\toprule
\textbf{Position Label} & \textbf{Example of Tweet}\\
\midrule
Pro & Should bring the death penalty back! $\#$executed \\\specialrule{\cmidrulewidth}{0pt}{0pt}
Con & The death penalty is pure violence, a barbaric and useless violence. \\\specialrule{\cmidrulewidth}{0pt}{0pt}
Neutral & I have such mixed feelings on the death penalty, some people deserve it but then some people don't $\#$executed \\
\bottomrule
\end{tabular}}
\centering
\caption{Example tweets of different position labels.}
\label{table:tweet_position}
\end{table*}

\begin{table}[ht]
\vspace{2ex}
\resizebox{\columnwidth}{!}{\begin{tabular}{ l | c | c | c }
\toprule
 & \textbf{Pro} & \textbf{Con} & \textbf{Neutral}\\
\midrule
Express emotion & 106 & 190 & 56 \\\specialrule{\cmidrulewidth}{0pt}{0pt}
Information sharing & 236 & 149 & 92\\\specialrule{\cmidrulewidth}{0pt}{0pt}
Social interaction & 44 & 84 & 43\\
\bottomrule
\end{tabular}}
\centering
\caption{Statistics for the \textit{Obamacare} data set.}
\label{table:obama_stats}
\end{table}

\begin{table}[ht]
\vspace{2ex}
\resizebox{\columnwidth}{!}{\begin{tabular}{ l | c | c | c }
\toprule
 & \textbf{Pro} & \textbf{Con} & \textbf{Neutral}\\
\midrule
Express emotion & 187 & 271 & 45 \\\specialrule{\cmidrulewidth}{0pt}{0pt}
Information sharing & 32 & 81 & 60\\\specialrule{\cmidrulewidth}{0pt}{0pt}
Social interaction & 144 & 131 & 49\\
\bottomrule
\end{tabular}}
\centering
\caption{Statistics for the \textit{Death Penalty} data set.}
\label{table:death_stats}
\end{table}

\begin{table}[ht]
\vspace{2ex}
\resizebox{\columnwidth}{!}{\begin{tabular}{ c | c | c | c | c | c | c}
\toprule
 & & \multicolumn{5}{c}{\textbf{Predictions}}\\
\midrule
\textbf{Model} & \textbf{Labelset} & $pp_1$ & $pp_2$ & $pp_3$ & $pt_1$ & $pt_2$\\
\midrule
$m_1$ & $\{pp_1,pt_1\}$ & $1$ & $-$ & $-$ & $0$ & $-$ \\
$m_2$ & $\{pp_3,pt_2\}$ & $-$ & $-$ & $0$ & $-$ & $1$ \\
$m_3$ & $\{pp_2,pt_2\}$ & $-$ & $0$ & $-$ & $-$ & $1$ \\
$m_4$ & $\{pp_1,pp_3\}$ & $0$ & $-$ & $1$ & $-$ & $-$ \\
$m_5$ & $\{pt_1,pt_2\}$ & $-$ & $-$ & $0$ & $1$ & $-$ \\
$m_6$ & $\{pp_1,pp_2\}$ & $1$ & $0$ & $-$ & $-$ & $-$ \\\midrule
\multicolumn{2}{c|}{average votes} & $2/3$ & $0$ & $1/3$ & $1/2$ & $1$\\
\multicolumn{2}{c|}{final predictions} & $1$ & $0$ & $0$ & $0$ & $1$\\
\bottomrule
\end{tabular}}
\centering
\caption{An example of RAkEL on purpose and position classification with $k = 2$ and $m = 6$.}
\label{table:rakel_example}
\end{table}

\section{Multi-Label Classification with Post-Processing}
\label{sec:multi_label}
\subsection{Random k-Labelsets for Multi-Label Classification}
\label{subsec:labelsets}
Different from traditional single-label classification task in which every instance is associated with only one single label, in multi-label classification, the instances are associated with a set of labels. In many application domains, multi-label cases are more common, for example, the movie \textit{The Lord of the Rings} can be classified into categories \textit{action}, \textit{adventure} and \textit{fantasy}\footnote{\url{http://www.imdb.com/title/tt0120737/}}. Therefore, multi-label classification has been a hot topic recently. In general, there are two types of multi-label classification methods: problem transformation and algorithm adaptation \cite{tsoumakas2007multi,tsoumakas2010random}. In problem transformation, methods transform the multi-label classification task into one or more single-label classification or ranking tasks. In algorithm adaptation, methods are extended in order to handle multi-label tasks directly. In this paper, we will use the method belonging to problem transformation.

Label powerset (LP) method \cite{boutell2004learning} is a simple but effective multi-label learning method which considers each unique set of labels that exists in the training set as one of the classes of a new single-label classification task and then the multi-label classification problem can be transformed into several single-label classification problems. 

RAkEL (Random k-Labelsets) multi-label classification method \cite{tsoumakas2010random} is based on LP. RAkEL solves the problems in label powerset (LP) method \cite{boutell2004learning} that the large number of labelsets when the number of labels is large and the inability to predict labelsets not observed in the training set while keeping the advantage of capturing label correlations. The RAkEL method breaks a large set of labels into a number of small-sized labelsets randomly, and for each of the labelsets, a multi-label classifier will be trained using LP method. 

For an unlabeled instance, the final decision is based on the combination of the results generated by all LP classifiers using the majority voting rule. In RAkEL, the size of labelset $k$ and the number of models $m$ can be specified. To utilize this model in our study, an example is shown in Table \ref{table:rakel_example} with $k = 2$ and $m = 6$. $pp_i$ denotes the $i^{th}$ purpose label and $pt_j$ denotes the $j^{th}$ position label. If the instance is predicted to be assigned this label, the value is $1$, otherwise the value is $0$. In this example, the threshold for the final prediction is $0.5$. The average vote is obtained by dividing the times of the label being predicted to be $1$ by the total number of this label being predicted. Therefore, for $pp_1$ and $pt_2$, both of the average votes, $2/3$ and $1$ respectively, exceed the threshold so the final prediction is $pp_1$ and $pt_2$.

\subsection{Post-processing Strategies for RAkEL}
\label{subsec:postprocessing}
In our application, it is assumed that each tweet has only one purpose label and one position label. However, multi-label classification methods consider all the labels equally. Therefore, some tweets may be predicted to contain no label or multiple labels for purpose or position if RAkEL method is used directly in our study. In order to deal with this problem, we propose the post-processing strategies for RAkEL method. For the tweets assigned no label or multiple labels for purpose and/or position, we will find \textit{K} tweets from the training set which are most similar to the original tweet and use the labels from these \textit{K} tweets to make new prediction. Two strategies can be used to make new prediction: ($1$) summation strategy; and ($2$) weighted summation strategy.

\begin{figure}[th]
\centering
\includegraphics[width=0.5\textwidth]{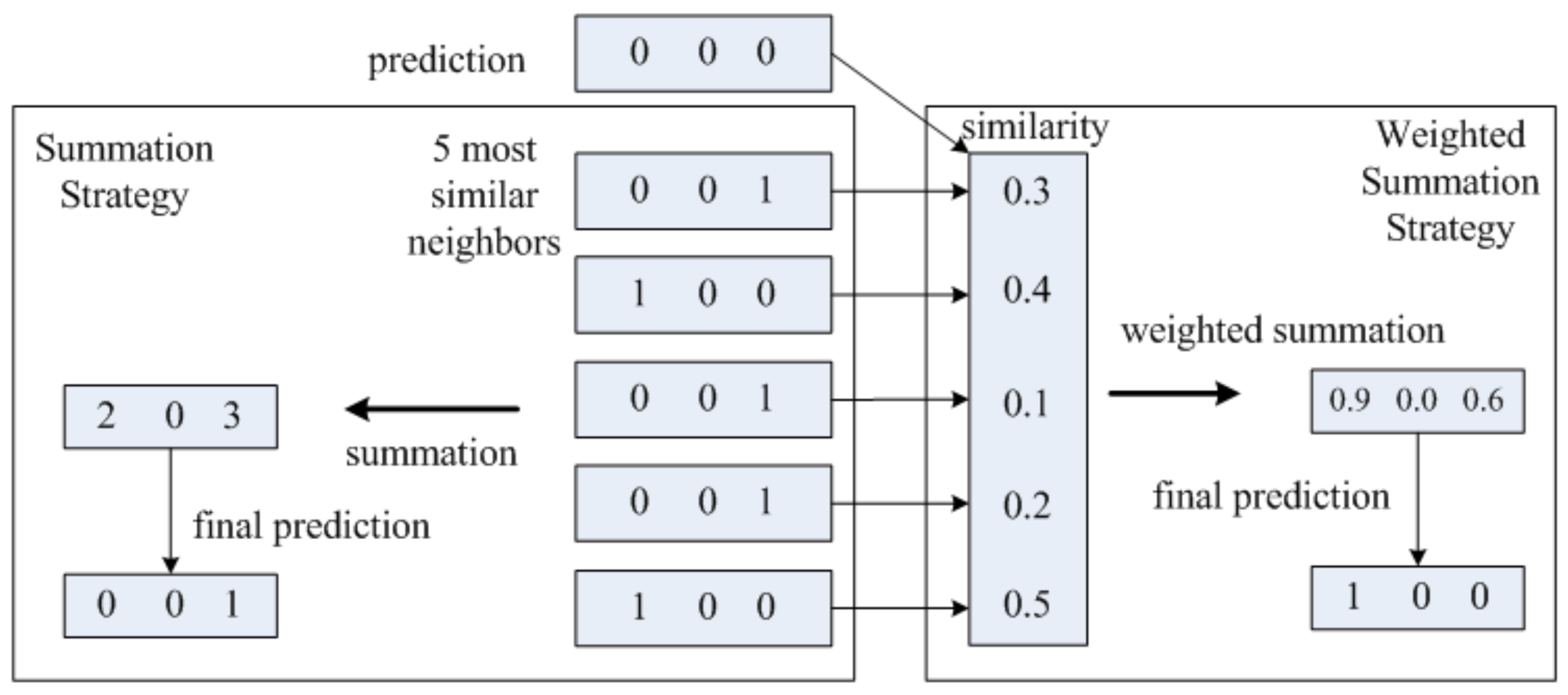}
\caption{The post-processing for multi-label classification.}
\label{fig:multi_post_processing}
\end{figure}

An example for the post-processing strategies is shown in Figure \ref{fig:multi_post_processing}. The original prediction for a tweet contains no labels, so we find $5$ most similar tweets for this tweet and the similarity between each neighbor and the original tweet are $0.3$, $0.4$, $0.1$, $0.2$ and $0.5$. In summation strategy, we sum up all the labels from $5$ neighbors and choose the index for the largest one as the new prediction for the original tweets, i.e., the third one with value $3$ in the figure. In weighted summation strategy, the label from each neighbor is first multiplied by the corresponding similarity value and then we sum them up to select the index for the largest one, i.e., the first one with value $0.9$ in the figure.

In order to find the most similar tweets, a similarity metric is required. In this study, we use the widely used cosine similarity metric which is a measure of similarity between two vectors of an inner product space that measures the cosine of the angle between them. Given feature vector $\mathbf{s}_i$ and $\mathbf{s}_j$, the cosine similarity $sim_{cosine}(\mathbf{s}_i, \mathbf{s}_j)$ is calculated as follows:

\begin{equation}
    sim_{cosine}(\mathbf{s}_i, \mathbf{s}_j) = \frac{\mathbf{s}_i \cdot \mathbf{s}_j}{\abs{\mathbf{s}_i} \times \abs{\mathbf{s}_j}}
\end{equation}

Combing RAkEL method and the post-processing strategies, the complete process is formed. It consists of three steps, i.e., training, testing and post-processing and the algorithm is presented in detail in Algorithm \ref{algo}.

\begin{algorithm}
\caption{Multi-label classification with post-processing}
\begin{algorithmic}
\REQUIRE \STATE Set of labels $L$ on size $M$, \\training set $D$, \\test set $T$, \\labelset size $k$, \\number of models $m$, \\number of most similar neighbors $K$, \\threshold $\epsilon$
\ENSURE \STATE Prediction results $Res$ \\
\STATE
\STATE
\texttt{// Step 1: Training process}
\STATE $S \leftarrow L^k$
\FOR{$i \leftarrow 1$ to $m$}
    \STATE Randomly select a labelset $r_i$ from $S$
    \STATE Train an $LP$ classifier $m_i$ based on $D$ and $r_i$
    \STATE $S \leftarrow S - \{r_i\}$
\ENDFOR \\
\STATE
\STATE 
\texttt{// Step 2: Testing process}
\FOR{each instance $x$ in test set $T$}
    \FOR{each label $j$ in $M$}
        \STATE $Sum_j \leftarrow 0$
        \STATE $Votes_j \leftarrow 0$
    \ENDFOR
    \FOR{each model $i$ in $m$}
        \FOR{labels $\lambda_j \in r_i$}
            \STATE $Sum_j \leftarrow Sum_j + m_i(x,\lambda_j)$
            \STATE $Votes_j \leftarrow Votes_j + 1$
        \ENDFOR
    \ENDFOR
    \FOR{each label $j$ in $M$}
        \IF{average vote $Avg_j > \epsilon$}
            \STATE $Res_j \leftarrow 1$
        \ELSE
            \STATE $Res_j \leftarrow 0$
        \ENDIF
    \ENDFOR
\ENDFOR \\
\STATE
\STATE 
\texttt{// Step 3: Post-processing}
\FOR{instance $x_t$ predicted without any label or with multiple labels in purpose and/or position}
    \STATE Select $K$ most similar neighbors from $D$
    \STATE Use strategy introduced in Section \ref{subsec:postprocessing} to make new prediction $Res_t$
\ENDFOR
\label{algo}
\end{algorithmic}
\end{algorithm}

\section{Experiments}
\label{sec:experiments}

\subsection{Experimental Setup}
\label{subsec:exp_setup}
In the experiments, we randomly choose $600$ tweets as the training set and the rest $400$ tweets as the test set for each data set. And we compare $5$ different methods in the experimental study including:

\begin{itemize}
    \item KNN: Since our proposed post-processing strategies are based on KNN model, we use KNN as one of the baselines in the comparison.
    \item SVM: Stated in Section \ref{subsec:labelsets}, the RAkEL method is based on LP method and LP method will use single-label classifers to make predictions. In the experiments, SVM is applied as the single-label classifier, so SVM is used as another baseline.
    \item RAkEL: The introduction of RAkEL is presented in Section \ref{subsec:labelsets}.
    \item RAkEL$+$sum: RAkEL with summation strategy.
    \item RAkEL$+$wsum: RAkEL with weighted summation strategy.
\end{itemize}

\subsection{Features}
\label{subsec:features}
In order to classify tweets, each tweet in the data set is represent as a vector of features and some commonly used text classification features are employed in the experiments including n-grams, punctuation, part-of-speech and Twitter-specific features. The details of features are shown below.

\begin{itemize}
    \item $n$-$gram$: We use the presence of n-gram, including $1$-$gram$ and $2$-$gram$ in the experiments, as the features, i.e., the value of this feature is $0$ or $1$. To reduce the dimensionality of $n$-$gram$ features, we remove the $1$-$grams$ occurring in the data set less than $2$ times and the $2$-$grams$ occurring in the data set less than $4$ times.
    \item punctuation: The number of occurrences of exclamation marks, question marks and colons.
    \item POS (part-of-speech): The number of occurrences of each POS tagger is used as the feature. The Tweet NLP and POS Tagging tool\footnote{\url{http://www.ark.cs.cmu.edu/TweetNLP/}} \cite{owoputi2013improved} is used to extract POS features for each tweet.
    \item Twitter-specific features: several typical Twitter-specific features are utilized including:
        \begin{itemize}
            \item the number of hashtags, i.e., the $\#$ symbol;
            \item the number of mentioning users, i.e., the $@$ symbol;
            \item the present of retweet, i.e., the $RT$ symbol;
            \item the number of hyperlinks including URLs and e-mail addresses.
        \end{itemize}
\end{itemize}

In the following experiments, we use $1$-$gram$, $2$-$gram$ and $(1$+$2)$-$gram$ to denote unigram, bigram and a combination of unigram and bigram features respectively. We also combine punctuation features and Twitter-specific features as the statistical features and use \textit{STAT} to denote this combination. \textit{POS} is used to denote all the POS features.

\begin{table*}[ht]
\vspace{2ex}
\begin{tabular}{ l | c | c | c | c | c | c | c | c | c | c }
\toprule
& \multicolumn{5}{c|}{\textbf{\textit{Obama care} data set}} & \multicolumn{5}{c}{\textbf{\textit{Death Penalty} data set}}\\\cmidrule{2-11}
& $f_1$ & $f_2$ & $f_3$ & $f_4$ & $f_5$ & $f_1$ & $f_2$ & $f_3$ & $f_4$ & $f_5$\\
\midrule
KNN & 0.6245 & 0.6667 & 0.6017 & 0.6025 & 0.5845 & 0.7125 & 0.7500 & 0.6874 & 0.6850 & 0.6536\\\specialrule{\cmidrulewidth}{1pt}{1pt}
SVM & 0.5575 & 0.5708 & 0.5575 & 0.5520 & 0.5342 & 0.6542 & 0.6608 & 0.6542 & 0.6450 & 0.6347\\\specialrule{\cmidrulewidth}{1pt}{1pt}
RAkEL & 0.4675 & 0.4983 & 0.4483 & 0.4517 & 0.4452 & 0.5992 & 0.5750 & 0.5988 & 0.5842 & 0.5725\\\specialrule{\cmidrulewidth}{1pt}{1pt}
RAkEL+sum & 0.4583 & 0.5025 & 0.4325 & 0.4315 & 0.4220 & 0.5975 & 0.5733 & 0.6017 & 0.5788 & 0.5650\\\specialrule{\cmidrulewidth}{1pt}{1pt}
RAkEL+wsum & 0.4571 & 0.5025 & 0.4310 & 0.4235 & 0.4112 & 0.5967 & 0.5733 & 0.6000 & 0.5742 & 0.5554\\
\bottomrule
\end{tabular}
\centering
\caption{The Hamming loss of different methods on \textit{Obama care} and \textit{Death Penalty} data set. The notations $f_1$, $f_2$, $f_3$, $f_4$, and $f_5$ denote features $1$-$gram$, $2$-$gram$, $(1+2)$-$gram$, $(1+2)$-$gram$ + $POS$ and $(1+2)$-$gram$ + $POS$ + $STAT$. Smaller value denotes better performance.}
\label{table:results}
\end{table*}

\subsection{Evaluation Metrics}
\label{subsec:eval_metrics}
Different evaluation metrics have been used in multi-label classification \cite{tsoumakas2007multi}. In this study, we apply Hamming loss to evaluate the performance. Hamming loss, based on Hamming distance, takes into account the prediction error (an incorrect label is predicted) and the missing error (a relevant label not predicted), normalized over total number of classes and total number of examples \cite{sorower2010literature}. The Hamming loss is defined as follows:

\begin{equation}
    Hamming\; Loss = \frac{1}{|D|} \sum\limits_{i=1}^{|D|} \frac{S_i \oplus Y_i}{|L|},
\end{equation}

where $\abs{D}$ is the number of examples in the test data and $\abs{L}$ is the number of labels. $S_i$ and $Y_i$ denote the sets of true and predicted labels for instance $i$ respectively. $\oplus$ stands for the symmetric difference of two sets and corresponds to the exclusive OR (\texttt{XOR}) operation in Boolean logic \cite{tsoumakas2007multi}. Intuitively, the performance is better, when the Hamming Loss is smaller. $0$ would be the ideal case indicating that there is no error in the prediction. 

\subsection{Experiment Results}
\label{subsec:exp_results}
Since multi-label classification method is employed in the experiments, the Hamming loss is applied as the evaluation measure. However, the single-label classifiers used in the experiments like KNN and SVM cannot be evaluated directly using Hamming loss. Therefore, the purpose labels and position labels generated by two individual classifiers are combined and the combined labels are in the same form of results generated by multi-label classifiers.

To validate the effectiveness of the multi-label classification method and the post-processing strategies in this application, the SVM method are used as the baselines. Two SVM classifiers are trained for purpose and position respectively and LIBSVM tools\footnote{\url{http://www.csie.ntu.edu.tw/~cjlin/libsvm/}} are used for training the SVM classifiers. Moreover, our proposed post-processing strategies are based on KNN method, so we also report the results generated by KNN in the experiments (the parameter \textit{K} is set to be 10). For the multi-label classification, we use the implementation of RAkEL in Mulan\footnote{\url{http://mulan.sourceforge.net/}}. The results of the experiments are shown in Table \ref{table:results}. In the table, $f_1$, $f_2$, $f_3$, $f_4$, and $f_5$ denote features $1$-$gram$, $2$-$gram$, $(1$+$2)$-$gram$, $(1$+$2)$-$gram$ + $POS$ and $(1$+$2)$-$gram$ + $POS$ + $STAT$, respectively. In these results, the number of similar neighbors \textit{K} is also set to be 10 and the influence of this parameter will be discussed in the next section.

Some conclusions can be drawn from the results reported above.
\begin{itemize}
    \item Multi-label classification method RAkEL, no matter with or without post-processing, can perform better than single-label classifiers, i.e., KNN and SVM in this study on both data sets. However, due to the different characteristics of different data sets, the scales for the Hamming loss are different. Among all the methods, RAkEL with weighted summation strategy performs best. For example, RAkEL+wsum is improved $29.65$\% compared with KNN on \textit{Obama care} data set.
    \item RAkEL with post-processing using either summation strategy or weighted summation strategy can generate better results than the original RAkEL method, which validates the effectiveness of our proposed post-processing strategies. And weighted summation strategy is more effective on both data sets. For example, RAkEL+wsum and RAkEL+sum can get $7.6$\% and $5.21$\% improvement on \textit{Obama care} data set, respectively.
    \item In all the results, we can observe that using the combination of $(1$+$2)$-$gram$, $POS$ and $STAT$ features performs best. From $1$-$gram$ features to $(1$+$2)$-$gram$, $POS$ and $STAT$ features, the more features are introduced, the better performance can be achieved generally. For example, compare the performance of feature $f_5$ and $f_1$, the improvement is $10.04$\% on \textit{Obama care} data set.
\end{itemize}

\subsection{Sensitivity Analysis for the Postprocessing Stage}
\label{subsec:sensitivity}
In the postprocessing strategies, there is one parameter, i.e., the number of most similar neighbor \textit{K}, which may influence the performance. Therefore, in this section, we study the influence of different values of \textit{K}. Considering the combination of $(1$+$2)$-$gram$, $POS$ and $STAT$ features can perform best in above experiments, we only study the influence of \textit{K} using this combination of features on \textit{Death Penalty} data set.

\begin{figure}[th]
\centering
\includegraphics[width=0.5\textwidth]{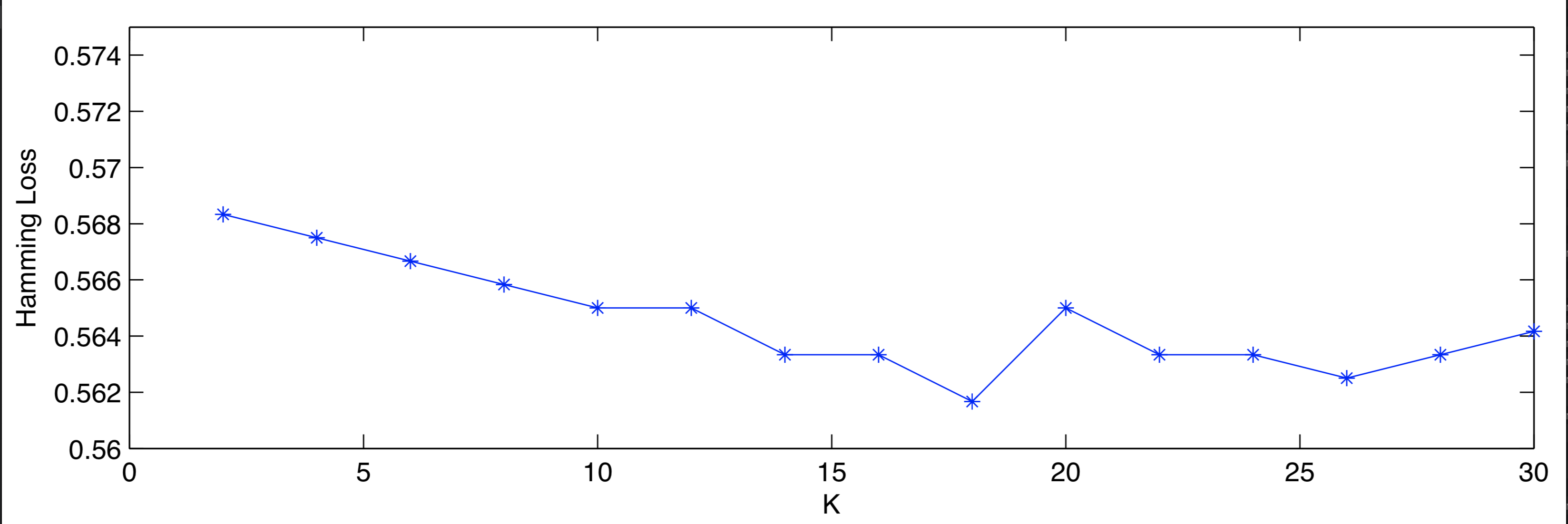}
\caption{The number of neighbors \textit{K} vs. Hamming loss using $(1$+$2)$-$gram$, $POS$ and $STAT$ features using RAkEL+sum strategy.}
\label{fig:rakel_sum}
\end{figure}

\begin{figure}[th]
\centering
\includegraphics[width=0.5\textwidth]{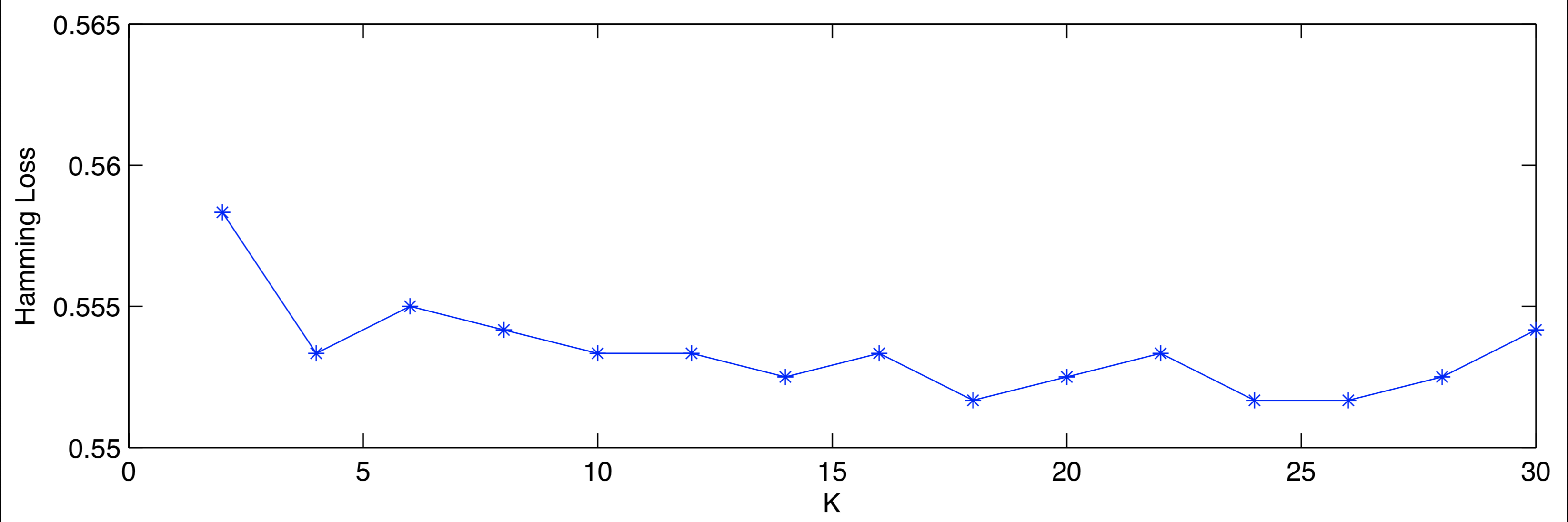}
\caption{The number of neighbors \textit{K} vs. Hamming loss using $(1$+$2)$-$gram$, $POS$ and $STAT$ features using RAkEL+wsum strategy.}
\label{fig:rakel_wsum}
\end{figure}

Using RAkEL+sum strategy and RAkEL+wsum strategy, the performance for different \textit{K} on $(1$+$2)$-$gram$, $POS$ and $STAT$ features are presented in Figure \ref{fig:rakel_sum} and Figure \ref{fig:rakel_wsum}. \textit{K} is set to range from $2$ to $30$ with the interval $2$. The results show that the best \textit{K}s for different post-processing strategies are different. For purpose classification, the optimal \textit{K} is around $18$ for RAkEL+sum strategy. But for RAkEL+wsum strategy, larger \textit{K} (around $24$-$26$) are preferred. However, since the differences in the results are all within $0.01$, the results are not very sensitive to the choice of \textit{K}.

\section{Conclusion and Future Work}
\label{sec:conclusions}
Analyzing purpose and position on tweets is beneficial for many areas. In this paper, we study the problem to identify tweet purpose and position simultaneously. We first transform this problem to a multi-label classification problem to capture the label correlations and then propose the post-processing strategies for a multi-label classification method RAkEL to classify tweets. To validate the effectiveness of our work, we build two data sets from Twitter related to the topic \textit{Obama care} and \textit{Death Penalty}. The experiments have been conducted on this data set and our results show that the proposed method outperforms the baseline method on accuracy. Furthermore, the influence of the parameters in the post-processing strategies has been studied.

In the future, we will further study this problem in two aspects. First, we will explore more features such as introducing negated context information and emotion lexicon. Secondly, we will integrate the specific constraints, i.e., each tweet can contain only one purpose label and one position label, in classification into the objective function in multi-label classification to form a unified framework for the classification task.

\bibliography{refs.bib}
\bibliographystyle{aaai.bst}

\end{document}